\documentclass{article-hermes}
\usepackage[pdftex]{graphics}
\usepackage{amsmath,amsfonts,amssymb,url}
\usepackage[all]{xy}

\title[La réduction de termes complexes]{La réduction de termes complexes dans les langues de spécialité}
\author{Yannis Haralambous* \andauthor Elisa Lavagnino**}
\address{* Enseignant-chercheur\\
Télécom Bretagne \& UMR CNRS 3192 Lab-STICC\\
** Doctorante de l'Université de Gênes et de Télécom Bretagne\\[6pt]
Technopôle Brest Iroise, CS 83818, 29238 Brest Cedex 3, France\\
\textup{\texttt{mailto:\{prénom.nom\}@telecom-bretagne.eu}}}

\resume{Nous étudions, par des méthodes statistiques sur des corpus français et italiens, le phénomène de réduction des termes complexes dans les langues de spécialité. Il existe deux types de réductions : anaphorique et lexicale. Nous montrons que la réduction anaphorique dépend du type de discours (de vulgarisation, pédagogique, spécialisé) mais ne dépend ni du domaine, ni de la langue, alors que la réduction lexicale dépend du domaine et est plus fréquente dans les domaines techniques à évolution rapide. D'autre part, nous montrons que la réduction anaphorique a tendance à suivre la forme pleine du terme, nous définissons une notion d'arbre anaphorique de terme et nous étudions ses propriétés. Concernant la réduction lexicale, nous tentons de démontrer statistiquement qu'il existe une notion de cycle de vie de terme, où la forme pleine est progressivement remplacée par une réduction lexicale.}

\abstract{Our study applies statistical methods to French and Italian corpora to examine the phenomenon of multi-word term reduction in specialty languages. There are two kinds of reduction: anaphoric and lexical. We show that anaphoric reduction depends on the discourse type (vulgarization, pedagogical, specialized) but is independent of both domain and language; that lexical reduction depends on domain and is more frequent in technical, rapidly evolving domains; and that anaphoric reductions tend to follow full terms rather than precede them. We define the notion of the anaphoric tree of the term and study its properties. Concerning lexical reduction, we attempt to prove statistically that there is a notion of term lifecycle, where the full form is progressively replaced by a lexical reduction.}

\motscles{Terminologie, langues de spécialité, extraction automatique de termes, termes complexes, réduction anaphorique, réduction lexicale.}
\keywords{Terminology, specialty languages, automatic term extraction, multi-word lexical units, anaphoric reduction, lexical reduction.}

\catcode`\@=11
\renewcommand{\submitted}[3][\today]{
   \renewcommand{\VAR@NomRevue}{#2}%
   \renewcommand{\VAR@NumeroSoumission}{%
      \ifthenelse{\equal{#3}{1}}{1\iere~}{#3\ieme~}}%
   \renewcommand{\VAR@SignatureArticle}{%
     \FonteSignature\raggedright
        Cet article a été publié dans la revue \emph{Traitement Automatique des Langues}, Volume~52/1, 2011, @ATALA. La publication originale est disponible sur le site \texttt{http://www.atala.org}.
   }
   \renewcommand{\VAR@EnTeteG}{\FonteEnTete
                               \textit{\VAR@NomRevue}\ 52/1}
   \renewcommand{\VAR@EnTeteD}{\FonteEnTete\VAR@TitreAbrege}
}

\begin{document}
\submitted{TAL}{1}
\maketitlepage
\def\trois{\up{$3+$}}

\section{Introduction}

En terminologie, nombreuses ont été les études portant sur les termes complexes  \cite{Collet2000,Jacques2004,Portelance1991}, leurs variantes réduites \cite{Freixa2002,Jacques2004,Moller1998}, les textes et le discours spécialisés et le mécanisme de la réduction dans les langues de spécialité  \cite{Jacques2005, Lavagnino2011}.

Pour justifier l'importance d'étudier le comportement des unités complexes dans le cadre des langues de spécialité, il suffit de dire qu'elles composent 80\% des éléments constituant les textes de spécialité \cite[p.~48]{Jacques2004}. Il est donc clair que leur rôle mérite d'être analysé et étudié dans tous ses aspects.

Notre contribution s'inscrit dans le cadre théorique de la \emph{terminologie textuelle et computationnelle} \cite{Bourigault1999,Conceicao2004} appliquée
aux discours spécialisés. Dans notre cas, le fait d'aborder les problèmes de la
variation des termes complexes en terminologie, c'est prendre en compte au
moins trois types de faits, à savoir :
\begin{itemize}\item les caractéristiques
de la structure interne d'un terme complexe \cite{Portelance1996,Collet2000}, son évolution \cite{Moller1998} et les effets que cette évolution peut avoir sur la langue de spécialité \cite{Jacques2005} ;

\item les caractéristiques des environnements
textuels étudiés \cite{Desmet2005,Alexandru2005}
l'usage et le comportement des
termes dans le discours \cite{Jacques2000} et les effets sur la communication, les
usagers et  la langue de spécialité \cite{Lerat2009,DeVecchi2008} ;

\item le traitement automatique des terminologies \cite{Daille1998} et la  terminologie computationnelle \cite{Savary2003}.
\end{itemize}

La nature du terme et surtout son
rôle dans le discours sont d'une importance fondamentale dans notre recherche.
Ce n'est que récemment que l'intérêt pour la variation des termes a été l'objet
de débats et d'analyses \cite{Soglia2002}. La fonction des termes n'est plus uniquement celle de
dénommer une entité dans un système conceptuel figé ; l'évolution continue
de la science et des techniques demande un renouvellement constant des
terminologies et leur adaptation aux nécessités des usagers et des situations
communicatives. 


\section{Autour de la définition du terme complexe}

Le terme complexe est une \emph{unité monoréférentielle} (critère de monoréférentialité) qui appartient à une \emph{langue de spécialité} (critère terminologique), et qui représente une \emph{notion univoque} (critère notionnel)~\cite{Collet2000}. La structure d'un terme complexe est \emph{binaire} par définition (critère de binarité : tête/expansion, déterminant/déterminé, hyper-/hyponyme, cf.\ ci-dessous). 

Parmi les critères structuraux définissant les termes complexes, la binarité occupe sans doute une place fondamentale puisque  tous les termes complexes ont une structure binaire\footnote{La binarité, en tant que caractéristique du terme complexe, nous permet de considérer sa structure divisée
en deux parties permettant une meilleure identification des rôles des
éléments : la tête, responsable de la caractérisation syntaxique du TC
et les constituants qui en définissent les changements au niveau
sémantiques. Pour l'instant, nous n'avons pas considéré les relations
hiérarchiques entre les éléments, qui pourraient nous permettre
d'envisager les rapports reliant les différentes composantes  du terme complexe \cite{Collet2000}.}, composée par deux éléments qui peuvent être simples ou composés \cite{Collet2000}. 

\begin{figure}[t]
\begin{minipage}{\textwidth}\fontsize{8pt}{10pt}\selectfont Il cambiamento climatico e il \textbf{degrado ambientale} sono suscettibili di provocare l'aumento della migrazione dall'Africa sub-sahariana, con effetti potenzialmente devastanti per le centinaia di milioni di persone, soprattutto povere, che lì vivono. [...] Oggi, il \textbf{degrado} è un problema serio per 32 Paesi dell'Africa, e oltre trecento milioni di persone che già affrontano la scarsità d'acqua. [...] Inoltre, in Senegal, le migrazioni interne e internazionali sono state provocate dai cambiamenti ambientali che hanno ridotto le opportunità di lavoro nell'agricoltura, la quale è diminuita con l'aumentare del \textbf{degrado ambientale}. [...] (\url{http://www.ecologiae.com/riscaldamento-globale-rifugiati-climatici-aumentare/14240/})

\medskip

I terreni vitati destinati alla produzione dei vini a \textbf{Denominazione di Origine Controllata} (D.O.C.) devono essere iscritti --- su denuncia dei conduttori interessati tramite il Comune territorialmente competente che opera la verifica attraverso l'Ispettorato Provinciale dell'Agricoltura --- in apposito Albo pubblico istituito presso ogni Camera di Commercio. [...] L'iscrizione all'Albo serve per poter effettuare la \og denuncia delle uve\fg{} --- da parte dei conduttori di terreni già iscritti - destinate alla produzione di vino a Denominazione di Origine Controllata, al Comune competente per territorio. Ciò al fine di commercializzare il prodotto con la rispettiva \textbf{denominazione}. [...] (\url{http://www.cameradicommercio.ag.it/index.php?option=com_content&task=view&id=47&Itemid=110})

\medskip

La \textbf{denominazione di origine controllata} \og{}Soave\fg{} e \og{}Soave classico\fg{} è riservata ai vini \og{}Soave\fg{} (anche con la specificazione della sottozona Colli Scaligeri), \og{}Soave\fg{} spumante e \og{}Soave classico\fg{}, che rispondono alle condizioni ed ai requisiti stabiliti dal presente disciplinare di produzione. [...] Le condizioni ambientali e di coltura dei vigneti destinati alla produzione dei vini a \textbf{denominazione controllata} \og{}Soave\fg{} e \og{}Soave\fg{} Classico devono essere quelle tradizionali della zona e, comunque, atte a conferire alle uve e al vino derivato le specifiche caratteristiche. [...] (\url{http://www.ilsoave.com/disciplinare.php})
\end{minipage}
\caption{Exemples de termes complexes dans leur contexte.\label{exemplesp4}}
\end{figure}

Par exemple (cf. fig.~\ref{exemplesp4}) :

\begin{itemize}
\item\emph{[degrado] [ambientale]} : le terme complexe est composé de deux constituants
simples. Ce facteur influence le mécanisme de la réduction qui, dans la plupart des termes à deux composants,  ne s'avère qu'en donnant lieu à des variantes anaphoriques, comme \emph{degrado}, ci-dessus.

\item\emph{[denominazione] [di] [origine controllata]} : le terme complexe est composé de deux constituants eux-mêmes composés. Cette structure permet la réalisation de deux typologies de réduction (variantes : \emph{denominazione} et \emph{denominazione controllata}, ou \emph{DOC}). Rappelons ici le debat concernant la relation entre le mécanisme de la siglaison et la réduction qui reste toujours ouvert. Dans notre étude, nous avons traité les deux mécanismes comme non équivalents à cause de la motivation qui les justifie: la siglaison n'est pas un mécanisme spontané repondant à des exigences textuelles, mais plutôt un mécanisme émanant des experts d'un certain domaine. \cite{Abreu1992}
\end{itemize}

La binarité peut etre analysée selon les trois niveaux structuraux des termes complexes :
\begin{itemize}\item une structure syntagmatique : tête et
composants ;
\item une structure sémantique :
déterminant/déterminé ;
\item
une structure onomasiologique :
hyperonyme/hyponyme.
\end{itemize}

Au niveau syntagmatique, le terme complexe est une unité composée linéaire qui est formée de composants considérés soit comme des éléments \emph{forts} (noms, verbes, adjectifs et adverbes) ou bien lexicaux, soit comme
des éléments \emph{faibles} (prépositions, articles et
conjonctions) ou bien grammaticaux \cite{Collet2000,Delmonte2004}.
Ces éléments sont en relation d'interdépendance grammaticale entre eux. 
La structure
syntagmatique du terme complexe peut être schématiquement représentée comme suit \cite{Collet2000} :
$$\text{terme complexe} =
\text{(tête)} + (\text{composant}_1 + \text{composant}_2 + \text{composant}_3).
$$

Lorsque nous nous situons au niveau
sémantique, nous retrouvons la tête (déterminé) et les composants (déterminants). Les déterminants caractérisent au niveau sémantique le déterminé, qui donc change son référent notionnel.  

Au niveau onomasiologique, les
composants sont étudiés selon la typologie des relations qui s'instaurent entre
eux, par exemple l'analyse des rapports d'hyperonymie/hyponymie. 

Pour représenter graphiquement cette
relation, prenons l'exemple \emph{metodo di produzione biologica}. Nous constatons que la tête \emph{metodo} instaure une relation d'hyperonymie/hyponymie avec le terme complexe en forme
pleine ; d'ailleurs nous distinguons plusieurs degrés de
dépendance :
$$
\begin{matrix}
\text{metodo} \xrightarrow{\text{dépendance}} \text{di produzione}\\
\downarrow\\
\text{metodo di produzione} \xrightarrow{\text{dépendance}} \text{biologica}\\
\downarrow\\
\text{metodo di produzione biologica}.
\end{matrix}
$$

La binarité d'un terme complexe est à la base de son instabilité et des changements que sa linéarité peut subir aux trois niveaux structuraux. Ces modifications sont determinées  par les déplacements des composants ou bien leur chute, comme dans le cas de la réduction. 

Pour l'exemple \og agricoltura biodinamica\fg{} nous avons :
\begin{itemize}
\item plan syntagmatique: $[\text{agricoltura}] (\text{tête}) + [\text{biodinamica}] (\text{composant})$ ;
\item plan sémantique:
$[\text{agricoltura}] (\text{déterminé}) + [\text{biodinamica}]
(\text{déterminant})$ ;
\item plan onomasiologique :
$[\text{agricoltura}] (\text{hyperonyme}) + [\text{biodinamica}] (\text{hyponyme})$.
\end{itemize}

Si le composant \emph{biodinamica} tombait, des changements seraient évidents à tous
les niveaux. Il est donc fondamental de souligner que même si la réduction
affecte tous les niveaux structuraux d'un terme complexe, les effets de ce mécanisme ne se
reflètent pas sur la fonction dénominative du terme complexe plein.

\section{Autour de la définition de la réduction}

Plusieurs études se sont intéressées
à la question fondamentale suivante : \og comment
l'immersion dans un discours permet-elle d'omettre une partie \emph{a priori} essentielle d'un terme complexe, c'est-à-dire formé par plusieurs mots, comme par exemple `contrôleur de carrefour', `équipement de terrain', `effort à la commande', `réseau routier national', etc. ?\fg{}  \cite{Jacques2005}.

La réduction est un mécanisme discursif
qui, à travers l'élision d'au moins un constituant, transforme globalement un
terme complexe en maintenant ses noyaux référentiel et notionnel
\cite{Collet1997}. 

Sur la base des recherches de
Guilbert \cite{Guilbert1975} et Portelance \cite{Portelance1989}, on peut
affirmer que la réduction coïncide avec la suppression d'une information \emph{non différentielle}, c'est-à-dire
l'élément ou les éléments non fondamentaux pour la catégorisation du terme.
Toutefois, cette acception limite ce mécanisme aux seuls phénomènes
sociolinguistiques, sans tenir compte du fait que la réduction peut répondre
également des exigences contextuelles
reliées au contexte et au cotexte \cite{Alexandru2005} ; cette acception
diminue, en outre, la valeur de la réduction et exclut la possibilité que
ce mécanisme puisse former des alternatives au syntagme plein. La réduction
représente en revanche, comme nous le verrons, un élément important de cohésion
textuelle. 

De manière générale, ce phénomène a
été traité soit comme un simple mécanisme d'anaphore qui permet par
exemple de reprendre \emph{le parc naturel} par \emph{le/ce parc}, soit
comme un processus de lexicalisation qui, par exemple, a transformé \emph{voiture
automobile} en \emph{automobile} ou \emph{téléphone portable} en \emph{portable}.
Schématiquement, l'effacement de l'expansion du terme complexe resterait, en
tant qu'anaphore, étroitement dépendant du contexte et du cotexte, tandis que l'effacement de
sa tête produirait une nouvelle unité lexicale. Le discours instaure ainsi un
jeu entre les phénomènes de réduction participant à sa propre cohésion et ceux
participant à la construction de nouvelles unités lexicales d'un domaine. 

En outre, la variation par réduction
satisfait le besoin d'économie de la langue \cite{Thi2008}. En effet, la répétition de la
forme pleine du syntagme devient lourde et non économique, alors que l'emploi d'une variante réduite conserve le noyau notionnel et référentiel, tout en évitant l'ambiguïté. Collet \cite{Collet2004} ajoute que la réduction, en ce
qu'elle est précisément due à la réitération du terme complexe, d'une part, et
en ce qu'elle constitue une forme différente de sa source, d'autre part, est à
ranger parmi les moyens de cohésion lexicale. La finalité de la réduction est
donc double : sur le plan de l'encodage \cite{Jacques2004}, elle permet
une économie d'énergie, sur le plan du décodage \cite[p. 528--529]{Collet2000}, elle
contribue à la cohésion textuelle en ce qu'elle constitue une forme de
répétition d'un item lexical. Les caractéristiques de ces microstructures (ce
que Collet appelle les \og contextes réductionnels\fg) permettent la réduction et
autorisent les locuteurs à se dispenser d'une partie \emph{a priori} essentielle du
terme. Ce mécanisme est donc en relation avec l'organisation plus globale du
texte, témoignage de l'interaction entre le lexique et le discours.

\subsection{La réduction à la base du polymorphisme}

À travers le mécanisme de la réduction, une notion représentée par un terme complexe se trouve être dénommée par plusieurs unités qui partagent le même référent dans un domaine de spécialité. Dans
les textes examinés, nous avons trouvé diverses formes du terme complexe comme dans l'exemple qui suit:
\begin{itemize}
\item terme plein : \emph{mode de production biologique} ;                
\item variante 1 : mode  ;
\item variante 2 : mode de production  ;
\item variante 3 : production biologique .  
\end{itemize}

Ainsi, certains termes complexes sont réalisés dans le même texte, sous
trois formes différentes :
\begin{itemize}
\item une forme que nous
appellerons \og pleine\fg, par exemple \emph{mode de production biologique},
\item une forme réduite à
la tête du terme, \emph{mode de production} ou \emph{mode},
\item pour certains, une
forme réduite à l'expansion du terme, e.g. \emph{production
biologique}.
\end{itemize}
 
La polymorphie désigne la coexistence de plusieurs formes possibles d'expression pour
certains termes complexes. S'il y a réduction, c'est parce qu'il y a adaptation
du producteur du texte aux nécessités du discours. Certaines de ces réductions
procurent un réel bénéfice sur le plan sémantique, en permettant de condenser
certaines informations sur une seule occurrence. Par exemple le terme \emph{produit issu de l'agriculture biologique} étant très long et peu
économique, les usagers se sont tournés vers une forme abrégée qui puisse
substituer le terme complexe dans le texte : \emph{produit biologique}.

Telle qu'illustrée ci-dessus, la coexistence de diverses formes d'un terme complexe
--- une forme pleine et une ou plusieurs formes réduites --- suscite
deux questions essentielles. La première concerne la compréhension des facteurs
de réduction et du mécanisme à l'\oe uvre dans l'effacement des constituants
majeurs d'un terme complexe ; il s'agit de comprendre pourquoi et comment, en
un point donné du discours, l'auteur choisit plutôt une forme qu'une autre, et
comprendre ce qui, dans le discours, l'autorise à user d'une forme dans
laquelle toute l'information attachée au terme complexe n'est pas exprimée.
Cela permet également d'expliquer ce qui rend possible et favorise l'effacement de constituants. La seconde question concerne les conséquences sémantiques de
l'effacement de constituants. L'enjeu
est de dégager les facteurs propres au discours qui permettent
de mieux comprendre la polymorphie des termes complexes.

\subsection{La valeur des variantes réduites}

Le terme complexe et les variantes correspondantes donnent vie à des \emph{couples réductionnels}. Les variantes
peuvent avoir différents degrés d'autonomie référentielle, selon leur charge
sémantique par rapport au terme complexe en forme pleine. La charge sémantique d'une
composante représente sa valeur informative à l'intérieur du terme : plus une
composante est fondamentale pour la transmission du sens du terme complexe, moins elle
subira l'effet de la réduction. La valeur informative d'une composante ne
constitue pas le seul facteur qui influence sa chute potentielle. L'usage et
son figement à l'intérieur d'une terminologie représentent d'autres facteurs
qui peuvent être cause de réduction. Comme déjà anticipé, même les exigences
textuelles peuvent être un facteur causant la réduction. 

Selon Guilbert \cite{Guilbert1975} et Portelance \cite{Portelance1989}, nous
observons que la réduction est un mécanisme qui supprime l'information non-différentielle, c'est-à-dire l'élément ou les éléments qui ne caractérisent
plus le terme complexe.

Le mécanisme de la réduction peut également être justifié par des raisons
pragmatiques : le facteur le plus incisif dans le cas de la réduction
anaphorique est l'organisation du discours ; contrairement à cela, la réduction
lexicale est influencée par des procédures lexicales qui donnent lieu a des
unités terminologiques indépendantes.

Nous pouvons déjà en déduire qu'il existe des conditions internes et
externes aux termes complexes, et que celles-ci peuvent influencer leur évolution. Les
conditions internes sont plutôt reliées aux caractéristiques du terme complexe et à sa
structure, tandis que les condition externes subissent l'influence de l'environnement
textuel, des caractéristiques de la langue de spécialité considérée et du degré
de figement des termes complexes. 

\begin{figure}[th]
\begin{minipage}{\textwidth}\fontsize{8pt}{10pt}\selectfont [...] Au niveau européen, les règlements (CEE) n\up{o} 2092/91 du 24 juin 1991 et (CE) n\up{o} 1804/99 du 19 juillet 1999 définissent les règles du mode de production biologique et du contrôle des produits, le premier pour les végétaux et le second pour les produits animaux. Les organismes génétiquement modifiés (OGM) et produits dérivés sont exclus des \textbf{modes de production biologique}.

\medskip

Fin 2004, les surfaces en \textbf{production biologique} (déjà certifiées ou en conversion) s'élèvent à 535 000 hectares, soit 1,9\% de la surface agricole utilisée (SAU) française.  [...] Ce \textbf{mode de production} touche environ 11\,000 exploitations, ce qui correspond à la moitié des 25\,000 exploitations et 1~million d'hectares visé par le plan pluriannuel de développement de l'agriculture biologique, présenté en décembre 1997 par le ministre en charge de l'Agriculture. [...] L'implantation de ce mode d'exploitation est faible dans les zones de grandes cultures intensives du Bassin parisien.
(\url{http://www.stats.environnement.developpement-durable.gouv.fr/donnees-essentielles/activites-humaines/agriculture-et-environnement/l-agriculture-biologique.html})

\medskip

Le vin bio c'est le sang de la terre, différent d'un terroir à l'autre, sain, authentique, sans artifices. C'est un nectar qui célèbre la vie, c'est l'invité incontournable des dîners festifs entre amis.
Pour obtenir du vin, le viticulteur qu'il soit en mode conventionnel ou en mode biologique doit cultiver sa vigne avant de vinifier son raisin. Nous allons vous exposer la spécificité du travail du viticulteur en mode viticulture biologique. [...] (\url{http://www.terroirselect.info/territoires/Provence/cultiver-vigne-mode-bio.html})

\medskip

\hrulefill

\medskip

[...] Il \textbf{metodo di produzione biologico} è disciplinato a livello comunitario dai reg. CE 2092/91 (normativa base) e 1804/99 (disposizioni per le produzioni animali). L'Italia è il primo paese in Europa per numero di aziende che applicano il metodo di produzione biologico. [...] (\url{http://www.coldiretti.it/anagribios/agricoltura.htm})

\medskip

[...] L'agricoltura biologica è un \textbf{metodo di produzione} definito dal punto di vista legislativo a livello comunitario con un regolamento, il Regolamento CEE 2092/91, e a livello nazionale con il D.M. 220/95. [...]
(\url{http://www.aiab.it/index.php?option=com_content&view=article&id=112&Itemid=136})

\medskip

[...] Parla di dati incoraggianti anche il sottosegretario alle Politiche agricole alimentari e forestali, Stefano Boco, che \og dimostrano una significativa attenzione degli agricoltori verso il \textbf{metodo biologico}\fg. [...]
(\url{http://www.aiol.it/last_node/articolo?page=863})
\end{minipage}
\caption{Exemples de réductions dans leur contexte.\label{exemplesp8}}
\end{figure}

Dans la section suivante, nous introduisons les deux types de réductions analysés dans notre étude.

\subsection{Les réductions anaphorique et lexicale}

En général, nous pouvons identifier
différents types de processus de réduction qui peuvent être classés sur un
continuum. Celui-ci est organisé selon les relations qui s'instaurent parmi
les termes complexes en forme pleine et les variantes réduites, les composantes qui chutent et la valeur de la variante qui se forme. De toute façon, dans toutes les études \cite{Collet2000,Jacques2004,Adelstein2002,Cardero2003} de ce mécanisme, ce sont les réductions anaphorique et lexicale qui retiennent l'attention de la plupart des auteurs.

La réduction anaphorique est un processus
discursif et textuel, tandis que la réduction lexicale est générée par des
conditions internes au syntagme plein (caractéristiques morphosyntaxiques,
notionnelles, statut terminologique des constituants) ou par des conditions externes
(niveau de spécialité du texte, typologie textuelle). Cette distinction est
globalement acceptée, mais certains spécialistes utilisent des dénominations
différentes pour les mêmes notions. Par exemple Aler-Dewolw \cite{Aler1986} parle d'\og ellipses contextuelle et néonymique\fg, Kocourek \cite{Kocourek1991} d'\og ellipses
contextuelle et lexicale\fg, Jacques \cite{Jacques2005} définit les variantes comme
\og reprise anaphorique\fg{} et \og terme réduit sans antécédents textuels\fg.
Dans cet article (ainsi que dans \cite{Lavagnino2011}) nous avons choisi les termes de \og réductions lexicale et anaphorique\fg, qui nous semblent les plus transparents.

Au niveau syntaxique phrastique et interphrastique, les variantes ont la même valeur. Nous détectons, par contre, une différence au niveau interne de la structure du terme complexe : une variante lexicale peut s'avérer sous la forme d'un changement non linéaire de la structure de terme complexe (par exemple : \emph{mode de production biologique} et \emph{mode biologique}), tandis que la réduction anaphorique entraîne un changement qui est toujours linéaire et qui détermine la chute des composants qui suivent la tête du terme complexe (par exemple \emph{degrado ambientale} et \emph{degrado}).

Afin de différencier les deux formes de réduction,
\cite{Jacques2000} affirme que :
\begin{itemize} 
\item la réduction lexicale n'est pas liée au milieu contextuel
  immédiat, qu'elle est durable, qu'elle crèe des variantes susceptibles de
  devenir des membres permanents de la terminologie du domaine, et qu'au
  niveau onomasiologique le syntagme plein est
  conservé même hors contexte ;
\item la réduction anaphorique a une valeur contextuelle, qu'elle se déroule au noyau-même du terme complexe, qu'elle a une valeur cohésive parmi les phrases d'un même texte et qu'au niveau onomasiologique, le syntagme plein
  est conservé uniquement en contexte.
\end{itemize}  

Pour mieux\enlargethispage{.5\baselineskip}
expliquer notre propos, voici quelques exemples (cf. fig.~\ref{exemplesp8}) :
{\footnotesize
\begin{center}
\begin{tabular}{|r|c|c|}\hline
&Exemple italien&Exemple français\\\hline
Terme&\emph{metodo di produzione biologica}&\emph{mode de production biologique}\\\hline
Syntagme&[metodo] subs. + [di] prép. +&[mode] subs. + [de] prép. +\\\relax
nominal plein&[produzione] subs. + [biologica] adj.& [production] subs.  + [biologique] adj.\\\hline
Tête&metodo&mode\\\hline
Constituants&di produzione biologica&de production biologique\\\hline
Réduction&metodo biologico&mode biologique\\
lexicale&&\\\hline
Réduction &metodo di produzione&mode de production\\
anaphorique&metodo&mode\\\hline
\end{tabular}
\end{center}

}

Dans
l'exemple italien, la tête du terme complexe reste liée à ses constituants dans le cas de
la réduction lexicale ; par contre pour la réduction anaphorique la
variante a seulement une valeur de cohésion anaphorique, alors que la tête
détient une charge sémantique supérieure.

À
travers ce type d'analyse, on définit un critère \emph{fonctionnel} : la variante
lexicalement réduite sert également comme connecteur textuel, mais elle ne se
limite pas à cette fonction. 

Du
point de vue onomasiologique, la différence entre les deux formes de
réduction réside dans le sens transmis aux variantes : dans le cas
de la réduction lexicale, la variante n'est pas un hyperonyme pur du terme complexe, mais elle en conserve la valeur onomasiologique ; au
contraire, la variante anaphorique représente l'hyperonyme du terme complexe, donc sur le plan onomasiologique
elle a moins d'intension et surtout est étroitement liée au contexte. Exemple:
{\footnotesize

\begin{center}
\begin{tabular}{|r|c|c|}\hline
Terme&\emph{logo comunitario di controllo CEE}\\\hline
Syntagme&[logo] subj. + [comunitario] adj. + [di] prép. + \\
 nominal plein&[controllo] subj. + [CEE]
acronyme/subst.\\\hline
Tête&logo\\\hline
Constituants&comunitario di controllo CEE\\\hline
Réduction&logo di controllo CEE,\\
lexicale& ou logo CEE (relation non hyperonymique)\\\hline
Réduction&logo (hyperonyme)\\
 anaphorique&\\\hline
\end{tabular}
\end{center}

}

Sur le plan \emph{formel}, on peut dire que la
réduction anaphorique permet seulement l'élision des constituants --- par contre
la réduction lexicale permet la suppression de la tête, des constituants, des
éléments forts et faibles. Exemple :
{\footnotesize

\begin{center}
\begin{tabular}{|r|c|c|}\hline
Terme&\emph{agricoltura biologica}\\
Syntagme nominal plein&[agricoltura] subst. +  [biologica] adj.\\
Tête&agricoltura\\
Constituants&biologica\\
Réduction lexicale&il biologico\\
Réduction anaphorique&agricoltura\\\hline
\end{tabular}
\end{center}

}
On a donc enregistré une élision du
constituant, dans le premier cas, et de la tête, dans le deuxième exemple. La
variante lexicale, qui détient une charge sémantique supérieure par rapport à
la variante anaphorique, peut substituer le terme complexe dans les textes, avec un risque
très réduit d'ambiguïté

Sur
le plan \emph{sémantique}, il y a également des différences. La variante
lexicale est caractérisée par une cohésion intérieure plus forte par rapport au
syntagme plein.

Par
exemple, le syntagme \emph{audit environnemental} se transforme en \emph{audit} :
il y a donc suppression de la cohésion interne du terme complexe. Mais, en cas de réduction lexicale, le comportement est à l'opposé : à
cause de l'élimination des constituants, la variante a une cohésion interne réduite,
voire parfois annulée. 

Ainsi, par
exemple, la variante lexicale \emph{metodo
biologico} du terme complexe \emph{metodo di produzione biologico} a une cohésion interne plus forte,
le signifié dans la forme réduite étant distribué parmi les constituants de la
variante, qui sont donc plus cohésifs et ont plus de valeur. Le signifié de la
forme pleine est distribué dans tous les éléments du syntagme.

En général, on peut affirmer qu'entre les variantes anaphoriques et les termes complexes, il s'instaure des relations de
type hyperonymique, et entre les variantes lexicales
et  les termes complexes des relations synonymiques, bien que ces variantes ne soient
pas de véritables synonymes.

\subsection{Réductions anaphorique et lexicale dans le discours}

Selon le domaine, le mécanisme de réduction peut se manifester de différentes manières. La
réduction anaphorique ne subit pas l'influence du domaine, vu qu'il s'agit d'un
phénomène intratextuel lié aux caractéristiques du
texte. En revanche, la réduction lexicale est influencée par le domaine. Plus
un domaine est technique et plus il sera caractérisé par une évolution rapide
de la terminologie qui peut produire \cite{Dury2010} :
\begin{enumerate}
\item l'effacement de certains éléments n'ayant plus une fonction différentielle dans un terme complexe ;
\item l'ajout d'éléments aux termes déjà existants ;
\item la disparition des termes.
\end{enumerate}
Dans le premier cas, nous retrouvons le mécanisme de la réduction.
Par contre, l'expansion des termes complexes peut, elle aussi, et dans un
deuxième temps, déterminer la réduction, si la linéarité du terme complexe change pour des raisons d'économicité.
Cette dernière affirmation est reliée au concept de \emph{série syntagmatique} :
pour chaque terme de la série qui enregistre un ajout d'information à travers
l'expansion de sa structure, nous pouvons constater l'effacement d'un
composant devenant implicite et d'apport non différentiel.

Ce mécanisme de réduction
assez fréquent s'appuie sur le principe de redondance et
s'apparente à l'apocope en langue parlée \cite[p.~103]{Portelance1991}.

En plus, un domaine à haut degré de réduction est normalement caractérisé
par des lexiques composés surtout par des termes complexes, et donc plus sensibles à subir
des mécanismes comme la réduction qui sont enchaînés dans des séries. Dans ses travaux, Portelance \cite{Portelance1989,Portelance1991,Portelance1996} décrit cette tendance des terminologies dans les domaines techniques, en justifiant la chute des éléments qui ont cessé d'être différentiels pour éviter leur redondance. Ce type de chute est à la base de la réduction lexicale dans le domaine technique. La réduction anaphorique dans les domaines techniques ne permet normalement pas l'évolution d'une langue de spécialité puisque les variantes anaphoriques ne sont pas indépendantes.

\section{Autour de la définition des corpus}

\subsection{Les langues de spécialité et les discours de spécialité}

~Si la langue de
spécialité est perçue comme une variété de la langue générale \cite{Prandi2006}, la variation des langues de spécialité est soumise aux mêmes
critères que la langue générale (variations diachroniques, diatopiques
et diastratiques). Nous pouvons retrouver d'autres
facteurs de variation qui influencent également la communication spécialisée,
comme la situation de communication, les intentions et les buts de la
communication. Ceux-ci conditionnent les ressources syntaxiques, morphologiques et textuelles utilisées dans les textes spécialisés (types de phrases, formes
verbales, articulateurs du discours, etc.) \cite{Desmet2001}.

Quand nous nous
référons aux langues de spécialité, il est important de ne pas confondre le
concept de discours avec celui de texte ou énoncé. Le discours se définit~
comme \emph{le produit des multiples pratiques discursives à l'\oe uvre dans la vie
sociale} \cite[p.~8]{Desmet2001}. Il ne peut pas
être dissocié du contexte socio-culturel auquel il
s'insère. L'énoncé \emph{représente sa manifestation ponctuelle, un
objet concret et observable}. Le texte enfin \emph{se réfère au modèle abstrait selon lequel s'organisent
les énoncés}.

Les textes peuvent être
classés selon des critères permettant d'identifier les niveaux 
suivants : 
\begin{enumerate}
\item niveau
fonctionnel (fonctions textuelles) ;
\item niveau situationnel (contexte
social des activités communicatives, lieu, temps, nombre, rôle et relations
entre les locuteurs) ;
\item
niveau du contenu sémantique (sujet d'un texte, différentes perspectives et développement thématique) ;
\item
niveau formel-grammatical (formes linguistiques et non linguistiques, aspects
grammaticaux, ressources syntaxiques et lexicales) \cite{Desmet2001}.
\end{enumerate}
Les critères que
nous venons de citer se distinguent en~:
\begin{itemize}
\item \emph{critères internes}, qui concernent directement la
structure du texte~;
\item
\emph{critères externes}, qui considèrent la contexte
où se situe le texte. 
\end{itemize}
Le
contexte est caractérisé par la situation communicative, l'émetteur, le
destinataire et les typologies textuelles. 

Un autre concept important à définir
est celui de domaine, \emph{seule façon de délimiter, de dénommer~ une structure cognitive, conceptuelle}
\cite[p.~187]{DeBesse2000}.

Notre projet se base sur deux approches différentes des
corpus~: d'une part, l'analyse linguistique du phénomène de la réduction s'appuie sur
une étude des corpus qui ont été constitués \emph{ad hoc}, donc selon une approche \emph{corpus-based} ; d'autre part, la validation
expérimentale est basée sur des corpus ayant été
constitués en vue, non pas de l'analyse de la réduction, mais de la validation des conclusions tirées de l'étude linguistique, donc selon une approche \emph{corpus-driven} \cite{Condamines2005}.

En parlant de \emph{corpus}, nous désignons l'aspect normatif
de la langue, notamment sa structure et son code. Le
corpus regroupe un ensemble de textes ayant une visée commune. Les
catégories des textes contenus dans un corpus peuvent être différentes, mais
normalement elles partagent un objectif commun. Les critères que nous considérons
comme les plus pertinents pour classer les textes dans un
corpus sont : le sujet, la perspective d'énonciation, le niveau de spécialisation, les sources, la typologie textuelle, la langue.

\subsection{Les typologies textuelles}

Le mécanisme de
réduction anaphorique dépend des caractéristiques du texte
dans lequel le terme est inséré. En général, nous pouvons relier ce mécanisme
au classement des typologies textuelles de Sabatini \cite{Sabatini,Lavagnino2011}. Plus un texte est contraignant et plus évidente
sera la tendance de la terminologie à rester figée, comme nous le montrerons
en section~\ref{secana}.

~
Cette classification est inspirée de celle de Desmet
\cite{Desmet2005} et vise à créer des typologies textuelles homogènes qui puissent être
utilisées dans les mêmes situations communicatives. Celles-ci, étant ciblées en langue de spécialité, imposent des contraintes
qui peuvent concerner le niveau situationnel influencé par les activités
communicatives, lieu, temps, nombre, rôle des locuteurs et relations entre eux.

Dans notre cas, les textes ont été classés selon des contraintes prenant en compte le niveau situationnel de la communication mais aussi le
contexte linguistique et le microcontexte (densité de
termes, structure du texte). 

Dans notre étude, nous avons traité les domaines indiqués dans le tableau~\ref{tabdomaines}. :
\begin{table}[ht]
\begin{center}
\begin{tabular}{|r|c|c|}\hline
Domaine&Réduction lexicale&Réduction anaphorique\\\hline
espaces naturels&oui&oui\\ 
médécine vétérinaire&non&oui\\
cancer&non&oui\\
emballages&oui&non\\ 
philosophie&oui&non\\\hline
\end{tabular}
\end{center}
\caption{Les domaines traités et les types de réduction étudiés.\label{tabdomaines}}
\end{table}

Pour chaque domaine, nous avons créé des corpus de référence, à partir du contenu textuel de revues spécialisées du domaine.

Pour ce qui concerne la réduction anaphorique, nous avons basé notre étude contrastive sur des textes tirés du Web et subdivisés  dans des catégories textuelles qui se basent sur la classification suivante :

\begin{itemize}
\item catégorie 1 : discours de vulgarisation et
de semivulgarisation scientifique ou technique. Par exemple, articles tirés des journaux généraux, brochures,
sites Web non spécialisés ;

\item catégorie 2 : discours scientifique ou technique à des fins pédagogiques. Par exemples textes universitaires, textes destinés aux experts des domaines pour la veille technologique et scientifique ;

\item catégorie 3 : discours scientifique ou technique spécialisé et/ou officiel, discours législatif. Par exemple, lois qui règlent les domaines, articles scientifiques.
\end{itemize}

\subsection{Statistique descriptive du comportement des termes complexes}

Plusieurs auteurs se sont intéressés à l'extraction de termes complexes dans les langues de spécialité \cite{Daille1996,Ananiadou1998,Jacquemin2001,SanJuan2002,NgongaNgomo:2010p3732}. La réduction, en tant que cas particulier de la variation, a également été étudiée \cite{Jacquemin1999,Daille2003,Nenadic2004}.

En marge de ces importants travaux, souvent liés à des développements d'outils d'extraction, il y a eu quelques études statistiques du comportement des termes, portant surtout sur la performance des outils \cite{Daille1998,Pazienza2005}.

Dans cet article nous nous intéressons spécifiquement au phénomène de réduction des termes complexes, notamment, nous visons à contribuer, par une meilleure compréhension du comportement des termes complexes, à l'optimisation des systèmes d'extraction de termes, de l'analyse sémantique et de la traduction automatique.

Notamment, nous avons distingué :
\begin{enumerate}
\item une phase concernant l'extraction automatique des termes complexes des textes de spécialité ;
\item une phase concernant la validation expérimentale des hypothèses qui sont décrites dans le chapitre précédent. 
\end{enumerate}
Les deux approches utilisent les mêmes protocoles qui visent des objectifs communs.

Pour ce qui concerne l'extraction automatique des termes, nous nous sommes servis du logiciel \emph{Acabit}. Cette validation empirique ne pouvait s'appuyer que sur des corpus créés \emph{ad hoc}, au risque de montrer les hypothèses de départ sans mettre en évidence d'éventuelles  contre-hypothèses. 
Pour éviter cela, et afin de vérifier ultérieurement les axiomes de départ, nous avons également décidé d'élaborer d'autres corpus qui se référaient à d'autres langues de spécialité. 

Dans le cas de la réduction anaphorique, nous avons constitué des corpus d'apprentissage d'où nous avons extrait les termes à travers \emph{Acabit}. Après avoir obtenu les listes de termes complexes associés à leurs variantes éventuelles, nous avons lancé des requêtes sur Internet pour retrouver d'autres textes. À ce stade, nous avons contacté des experts afin de procéder à la validation des listes des termes complexes et des variantes. La collaboration avec les experts s'est avérée fondamentale pour arriver à des conclusions sur les cas ambigus où les variantes anaphoriques pouvaient être confondues avec des hyperonymes du terme complexe.

En outre, pendant que les experts se concentraient sur la désambiguïsion des variantes, nous avons créé un instrument pour la catégorisation des textes. Cette plateforme informatique présente les textes classés selon le terme complexe détecté par \emph{Acabit} suivi d'une série de variantes éventuelles. 
Dans le cas de la réduction lexicale, le premier problème que nous avons rencontré concernait les typologies de termes complexes qu'il fallait traiter pour la validation des résultats. Nous avons décidé de traiter des termes complexes composés d'au moins 3 composants, considérés comme pertinents dans le cadre de notre étude.

Dans ce cas, les experts ont été contactés, en premier lieu, pour la confirmation des relations réductionnelles entre les termes complexes et les variantes détéctées, comme pour la réduction anaphorique. Ensuite, leur avis a été nécessaire pour une évaluation des composants qui subissaient le mécanisme de réduction. Cette deuxième problématique dérivait de l'analyse des langues de spécialité moins techniques, notamment par les composants adverbiaux. Leur effacement ne détermine pas de réduction, puisque la présence des adverbes détient une fonction de modulation de la valeur sémantique exprimée. Leur chute change l'inténsité du concept exprimé, par exemple \emph{dimensione propriamente etica} et \emph{dimensione etica}.

Concernant le choix des experts des domaines, nous nous sommes adressés à des experts différents selon la spécialité. Pour ce qui concerne le domaine des espaces naturels, nous avons contacté des organismes de recherche responsables de la base de données multilingue concernant les réseaux naturels transalpins. 
Pour le domaine vétérinaire, nous nous sommes adressés aux organismes de contrôles des activités agricoles biologiques, notamment ceux qui évaluent les élevages.  Ces organismes avaient déjà été consultés dans le cadre d'un projet concernant un glossaire multilingue sur les activités biologiques. Ils ont cumulé une expérience décennale dans l'évaluation des entreprises du domaine au niveau européen.

Pour ce qui concerne l'évaluation des termes appartenant au domaine de la philosophie, nous  avons contacté des chercheurs universitaires qui participent activement à la recherche dans les domaines spécialisés en philosophie et sciences humaines.

Enfin, dans le cadre de la médicine, nous avons collaboré avec les associations de divulgation de l'information en médicine citées dans les sources consultées pour la création de nos corpus. D'autre part, dans le cas des emballages, nous avons contacté les entreprises citées dans les revues prises en considération dans nos corpus pour avoir leur avis sur la terminologie utilisée.

\section{La réduction anaphorique}\label{secana}

\subsection{Hypothèses}

Les hypothèses que nous avons décidé de valider au niveau informatique sont
le résultat de réflexions concernant les facteurs qui nous venons de décrire au
niveau linguistique. 
Ainsi, nous formulons les hypothèses suivantes :

\begin{enumerate}
\item[\rm(1)] \emph{La réduction anaphorique est corrélée avec le type de discours, selon la classification donnée ci-dessus. En particulier, elle 
est plus présente dans les textes de catégorie 1, moins présente dans ceux de catégorie 2, quasi-absente dans ceux de catégorie~3.}
\item[\rm(2)] \emph{La propriété {\rm(1)} est indépendante du domaine et de la langue.}
\item[\rm(3)] \emph{S'agissant d'un phénomène anaphorique, dans un document les formes pleines ont tendance à apparaître avant les formes réduites (qui se réfèrent à elles).}
\end{enumerate}

\subsection{Validation expérimentale}

\subsubsection{Protocole utilisé} La première hypothèse est liée à des propriétés internes au texte. Pour la valider nous avons étudié les occurrences d'un certain nombre de termes complexes et de leurs réductions anaphoriques dans un corpus composé de documents tirés du Web.

\begin{figure}[t]
\resizebox{\textwidth}{!}{\includegraphics{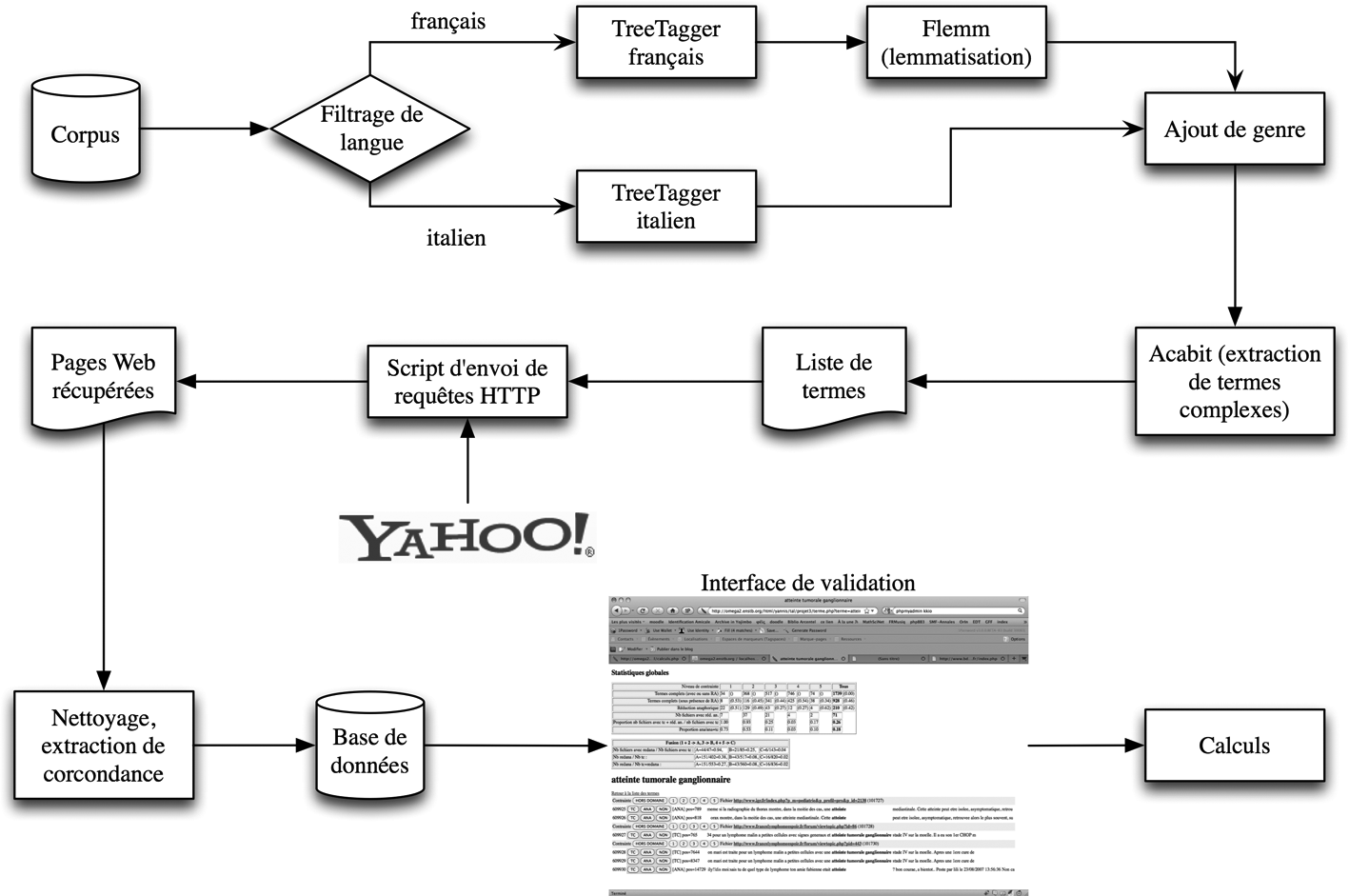}}
\caption{Le protocole utilisé pour la validation des hypothèses de réduction anaphorique.\label{diagramme}}
\end{figure}

Voici les cinq étapes du protocole utilisé (cf. fig. \ref{diagramme}) :
\begin{enumerate}
\item choix de termes complexes dans un domaine précis de langue de spécialité (cf.~\S\ref{acabit}) ;
\item récupération des documents contenant la forme pleine de chaque terme (cf.~\S\ref{recup}) ;
\item extraction des formes pleines ainsi que réduites (potentielles) des termes dans ces documents ;
\item validation par un expert de l'appartenance de chaque document au domaine, et balisage des formes réduites relevant de la réduction anaphorique ;
\item calculs.
\end{enumerate}

Les experts ont été sélectionnés selon les domaines de spécialité. Leur contribution a été assistée par des terminologues qui pouvaient resourdre les problèmes au niveau épistémologique. 

\subsubsection{Choix des termes complexes}\label{acabit} Pour obtenir une liste pertinente de termes complexes dans des domaines donnés (point 1 du protocole), nous avons choisi comme point de départ le contenu textuel de quatre revues spécialisées\footnote{\emph{Bulletin du cancer}
\url{http://www.john-libbey-eurotext.fr/fr/revues/medecine/bdc/sommaire.md}, ISSN 1769-6917 ; \emph{Vet.journal} \url{http://www.evsrl.it/vet.journal/}, sans ISSN ; \emph{Espaces naturels} \url{http://www.espaces-naturels.fr/a_la_une/la_revue_espaces_naturels}, ISSN 1637-9896 ;
\emph{Parchi} \url{http://www.parks.it/federparchi/rivista/}, sans ISSN.}, cf.~tableau~\ref{tabrevues}.
\begin{table}[ht]\small
\begin{center}
\begin{tabular}{|l|c|c|c|}\hline
Corpus & Langue & Années& Taille en Mo \\\hline
\emph{Bulletin du cancer} & français &1997--2010&43,02 Mo\\
Journal vétérinaire \emph{vet.journal} & italien &2003--2010&11,41 Mo\\
\emph{Espaces naturels} & français &2003--2009&\ 5,49 Mo\\
\emph{Parchi} & italien &1990--2009&\phantom{,49}18 Mo\\\hline
\end{tabular}
\end{center}
\caption{Les revues spécialisées utilisées comme point de départ pour obtenir des listes de termes complexes pour chaque domaine.\label{tabrevues}}

\end{table}
Comme on peut le voir sur la fig. \ref{diagramme}, l'extraction de termes complexes a été faite par le logiciel Acabit \cite{Daille1996} après balisage POS par TreeTagger et lemmatisation par Flemm (pour le français uniquement). Ces deux logiciels ne fournissant pas l'information du genre des noms, nous avons complété la chaîne de traitement en nous servant de ressources lexicales\footnote{Lexique 3 pour le français (\url{http://www.lexique.org/}), Morph-it! pour l'italien (\url{http://dev.sslmit.unibo.it/linguistics/morph-it.php}).} pour introduire cette information, qui améliore l'extraction de termes complexes en permettant la vérification de l'accord en genre.

\subsubsection{Récupération des documents Web contenant les termes}\label{recup} Le moteur de recherche Google n'autorisant pas la récupération de plus de 64 URL par requête, nous nous sommes tournés vers la plate-forme Yahoo BOSS, qui ne pose pas de limite sur le nombre de requêtes. Pour chaque requête Web, Yahoo retourne les mille premières URL, par ordre de pertinence. Nous avons récupéré tous les documents proposés par Yahoo, qui étaient récupérables et de format HTML, PDF ou texte brut, cf.~tableau~\ref{tabdocs}.
\begin{table}[ht]\small
\begin{center}
\begin{tabular}{|l|c|cccc|c|}\hline
Domaine & Langue & Nb doc., dont : & HTML & PDF & TXT & Taille totale\\\hline
Vétérinaire & italien & 14\,183 & 10\,420 & 3\,743 & 20 & 387 Mo\\
Cancer & français & 20\,790 & 13\,627 & 7\,124 & 39 & 647 Mo\\
Parcs nationaux & italien & 38\,502 & 28\,253 & 10\,226 & 23 & 1,15 Go\\
Parcs nationaux & français & 18\,200 & 13\,840 & 4\,328 & 32 & 849 Mo\\\hline
\end{tabular}
\end{center}
\caption{Les documents récupérés sur le Web.\label{tabdocs}}

\end{table}
Dans les chiffres ci-dessus, nous entendons par \og taille totale\fg{} celle des contenus textuels bruts obtenus après conversion et/ou nettoyage.
\subsection{Extraction des termes et validation par un expert} Par le biais d'une interface Web spéciale, reliée à une base MySQL, des experts ont pu (a) valider l'appartenance effective de chaque document au domaine donné, (b) indiquer le type du document (1 = vulgarisation, 2 = pédagogique, 3 = spécialisé) (c) vérifier les conditions de forme pleine ou de forme réduite anaphorique de chaque terme complexe observé.

\subsubsection{Arbre anaphorique d'un terme complexe}

Soit $T=(c_*)$ un texte (où $c_*$ sont des caractères), $t_*$ les occurrences de la forme pleine d'un terme complexe dans $T$, et $r_*$ celles des différentes formes réduites du même terme. L'ordre linéaire des caractères du texte induit un ordre des $t_*$ et des $r_*$. 

Notons $r_{i,1},\ldots,r_{i,n_i}$ les formes réduites placées entre les formes pleines $t_i$ et $t_{i+1}$ (ou entre $t_i$ et la fin du document). Dans le cas où il n'y a pas de tel $r_*$, on dira que $n_i=0$. Soit $\mathrm{pos}:T\to\mathbb{N}$ la fonction qui associe à chaque mot sa \og position\fg{} dans le texte (on compte le nombre de caractères depuis le début du fichier).

Dans $T$ il peut y avoir également des formes réduites $r'_*$ placées \emph{avant} la première forme pleine (c'est-à-dire telles que $\mathrm{pos}(r'_*)<\mathrm{pos}(t_1)$). Celles-ci sont appelées \emph{réductions cataphoriques} et on notera $r'_j$ la $j$-ème réduction cataphorique de la forme pleine $t_1$ (pour les autres $t_i$ on considère, dans ce modèle\footnote{On peut imaginer d'autres modélisations de ce phénomène. Ainsi, par exemple, on pourrait \og attacher\fg{} chaque réduction à la forme pleine la plus proche, que ce soit avant ou après elle. On aurait alors des réductions anaphoriques \emph{et} cataphoriques dans tout le document. Dans cet article, nous avons choisi de n'avoir de réduction cataphorique qu'avant la première forme pleine. Cf. aussi \S\,\ref{conclusions}, point 4.}, que l'on n'a que des réductions anaphoriques).

\begin{figure}[ht]
\begin{center}\resizebox{\textwidth}{!}{\includegraphics{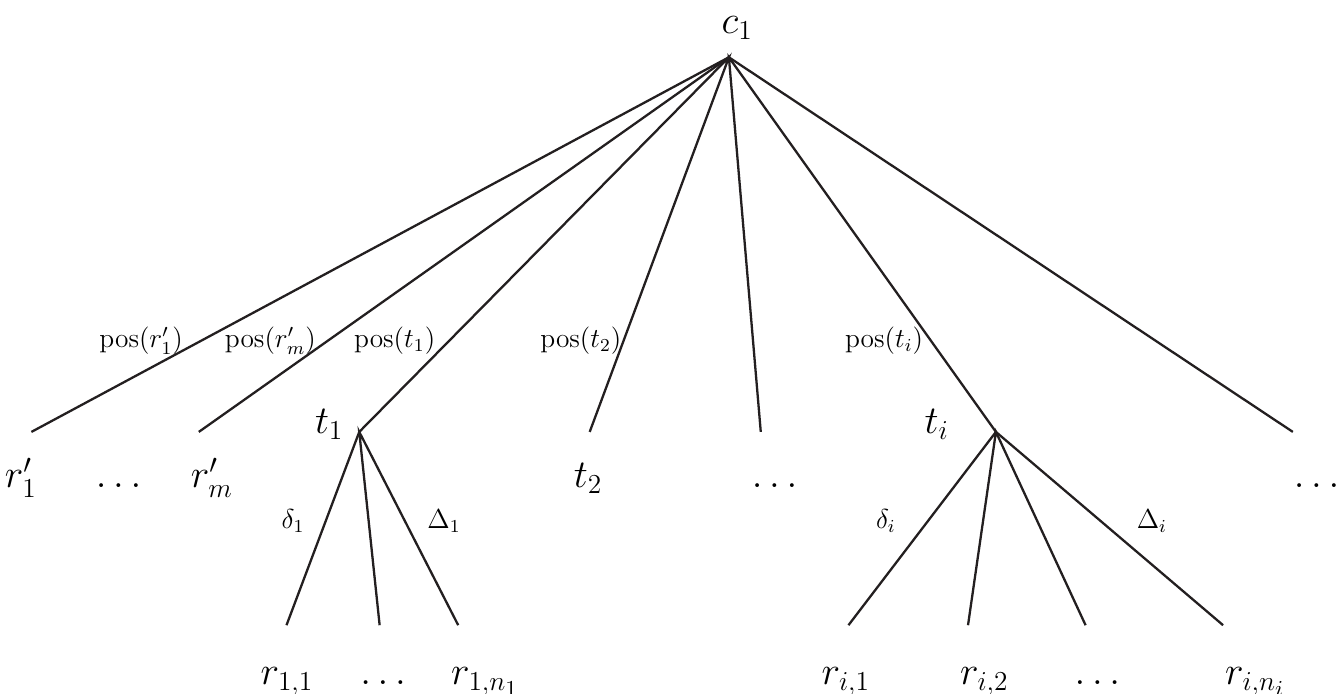}}\end{center}
\caption{L'arbre des réductions anaphoriques et cataphoriques d'un terme $t$ dont les $t_*$ (resp. $r_{*,*}, r'_*$) sont des occurrences de forme pleine (resp. de forme réduite).\label{arbre}}
\end{figure}

Les occurrences $t_*$, $r_{*,*}$ et $r'_*$ forment une structure d'arbre  ordonné et pondéré par la fonction $\mathrm{pos}$ (cf. fig. \ref{arbre}), que nous appelons \emph{arbre anaphorique du terme $t$}.

Pour étudier cette structure et en tirer des renseignements sur le comportement des réductions anaphoriques de $T$, nous allons nous intéresser à deux types de quantités :
\begin{enumerate}
\item des quantités relatives à la structure de l'arbre :
\begin{enumerate}
\item  
$d_m$, le degré moyen des n\oe uds $t_i$,
\item  
$d_-$, le nombre de réductions cataphoriques,
\item $f$, la moyenne des $f_i=\#\{t_j,t_{j+1},\ldots,t_k\mid d(t_{j-1})>0, d(t_{\ell})=0 \text{ pour } i\leq\ell<k \text{ et } d(t_{k})>0\}$ (où $d$ est le degré), c'est-à-dire le nombre de formes pleines consécutives entre deux formes réduites ;
\end{enumerate}
\item des quantités relatives à la pondération de l'arbre :
\begin{enumerate}
\item  
$\delta$, la moyenne des $\delta_i=\mathrm{pos}(r_{i,1})-\mathrm{pos}(t_i)$, c'est-à-dire la distance moyenne entre une forme pleine et sa première réduction anaphorique,
\item  
$\Delta$, la moyenne des $\Delta_i=\mathrm{pos}(r_{i,n_i})-\mathrm{pos}(t_i)$, c'est-à-dire la distance moyenne entre une forme pleine et la dernière réduction anaphorique avant la prochaine forme pleine ou avant la fin du fichier,
\item  
$\delta_-=\mathrm{pos}(t_1)-\mathrm{pos}(r'_m)$, c'est-à-dire la distance entre la dernière réduction cataphorique et la première forme pleine,
\item  
$\Delta_-=\mathrm{pos}(t_1)-\mathrm{pos}(r'_1)$, c'est-à-dire la distance la première réduction cataphorique et la première forme pleine ;
\end{enumerate}
\end{enumerate}

Le scénario suivant fournit une interprétation possible de ces quantités : après avoir écrit (en moyenne) $f$ formes pleines, l'auteur se sert (en moyenne) de $d_m$ formes réduites, dont la première arrive (en moyenne) à $\delta$ caractères après la dernière forme pleine. Éloigné de $\Delta$ caractères (en moyenne) de la forme pleine, il considère que l'ambiguïté est devenue trop forte, il revient à la forme pleine, et le cycle reprend.

De même, pour le cas cataphorique, l'auteur écrit en moyenne $d_-$ formes réduites cataphoriques avant la première forme pleine, qui est à une distance de $\Delta_-$ caractères de la première et de $\delta_-$ caractères de la dernière forme cataphorique.

\subsubsection{Première et deuxième hypothèses} 
Notre première hypothèse stipule que la quantité et le comportement des réductions anaphoriques et cataphoriques dépendent de la typologie de texte. Nous avons classé les documents en trois catégories (vulgarisation / pédagogique / spécialisé). Le lecteur trouvera dans le tableau~\ref{tabres} les moyennes $\bar\delta, \bar\Delta, \bar{d}_m, \bar{\delta}_-, \bar{\Delta}_-, \bar{d}_-, \bar{f}$ de $\delta$, $\Delta$, $d_m$, $\delta_-$, $\Delta_-$, $d_-$ et $f$ pour chacun des quatre corpus, ventilés par typologie de texte.
\begin{table}[t]\small
\begin{center}
\begin{tabular}{|r|ccc|ccc|ccc|ccc|}\hline
&\multicolumn{3}{c|}{Vétérinaire (italien)}&\multicolumn{3}{c|}{Cancer (français)}\\\hline
&vulg.&pédag.&spécial.&vulg.&pédag.&spécial.\\\hline
FP&3.20 & 4.09 & 5.16&4.97 & 10.96 & 11.66\\
ANA/FP&0.8 & 0.56 & 0.06&0.79 & 0.7 & 0.25\\
CATA/FP&0.13 & 0.01 & 0.00&0.12 & 0.02 & 0.00\\\hline
$\bar\delta$&3297.16 & 2074.05 & 3762.68&1917.76 & 2158.51 & 8118.12\\
$\bar\Delta$&6579.31 & 6797.56 & 4673.45&4485.39 & 10484.02 & 22292.91\\
$\bar{d}_m$&2.58 & 2.89 & 1.57&2.23 & 3.17 & 5.85\\
$\bar{\delta}_-$&2129.82 & 688.67 & NA&3043.83 & 1022.45 & NA\\
$\bar{\Delta}_-$&2805.29 & 688.67 & NA&6673.67 & 4907.64 & NA\\
$\bar{d}_-$&1.75 & 1.00 & NA&1.50 & 1.73 & NA\\
$\bar f$&1.75 & 2.42 & 3.39&1.72 & 1.99 & 7.38\\\hline
\end{tabular}

\medskip

\begin{tabular}{|r|ccc|ccc|ccc|ccc|}\hline
&\multicolumn{3}{c|}{Parcs (italien)}&\multicolumn{3}{c|}{Parcs (français)}\\\hline
&vulg.&pédag.&spécial.&vulg.&pédag.&spécial.\\\hline
FP&4.58 & 10.62 & 12.04&17.89 & 35.75 & 52.81\\
ANA/FP&0.73 & 0.55 & 0.19&1.03 & 0.50 & 0.17\\
CATA/FP&1.05 & 0.34 & 0.04&3.56 & 0.98 & 0.13\\\hline
$\bar\delta$&5233.06 & 2293.71 & 849.67&1773.01 & 972.44 & 993.31\\
$\bar\Delta$&7499.35 & 2832.36 & 10789.33&5964.44 & 1553.90 & 3817.10\\
$\bar{d}_m$&2.42 & 1.93 & 3.33&3.78 & 1.24 & 2.33\\
$\bar{\delta}_-$&1896.54 & 9931.29 & 2433.00&3340.20 & NA & NA\\
$\bar{\Delta}_-$&5216.58 & 13449.50 & 4811.00&13454.80 & NA & NA\\
$\bar{d}_-$&3.46 & 1.71 & 3.00&5.40 & NA & NA\\
$\bar{f}$&2.41 & 3.81 & 3.17&3.35 & 3.47 & 7.25\\\hline
\end{tabular}
\end{center}
\caption{Résultats de l'analyse des documents récupérés sur le Web. FP, ANA et CATA sont les nombres moyens de formes pleines (resp. anaphoriques, cataphoriques) par 100\,ko de texte, et ANA/FP et CATA/FP les ratios entre formes anaphoriques (resp. cataphoriques) et formes pleines. NA signifie que la donnée n'est pas calculable, faute de cas attestés dans les corpus.\label{tabres}}
\end{table}

De ces tableaux nous tirons les conclusions suivantes :
\begin{enumerate}
\item comme le montre la ligne FP, la densité de formes pleines est croissante lorsque le niveau de contrainte augmente (en allant de la vulgarisation au texte scientifique spécialisé). En effet, un texte scientifique évitera l'ambiguïté en favorisant les formes pleines, au détriment de l'économie ;
\item les lignes ANA/FP et CATA/FP montrent clairement que le nombre de réductions baisse lorsque l'on passe de la catégorie 1 aux catégories 2 et 3. C'est ainsi qu'est démontrée la première hypothèse ;
\item ce comportement est similaire dans les quatre corpus, appartenant à des domaines et à des langues différents, ce qui confirme la deuxième hypothèse ;
\item en ce qui concerne les autres paramètres, on ne peut en tirer aucune conclusion, puisqu'on ne constate aucune régularité significative.
\end{enumerate}

\begin{figure}[ht]
\resizebox{1.03\textwidth}{!}{\includegraphics{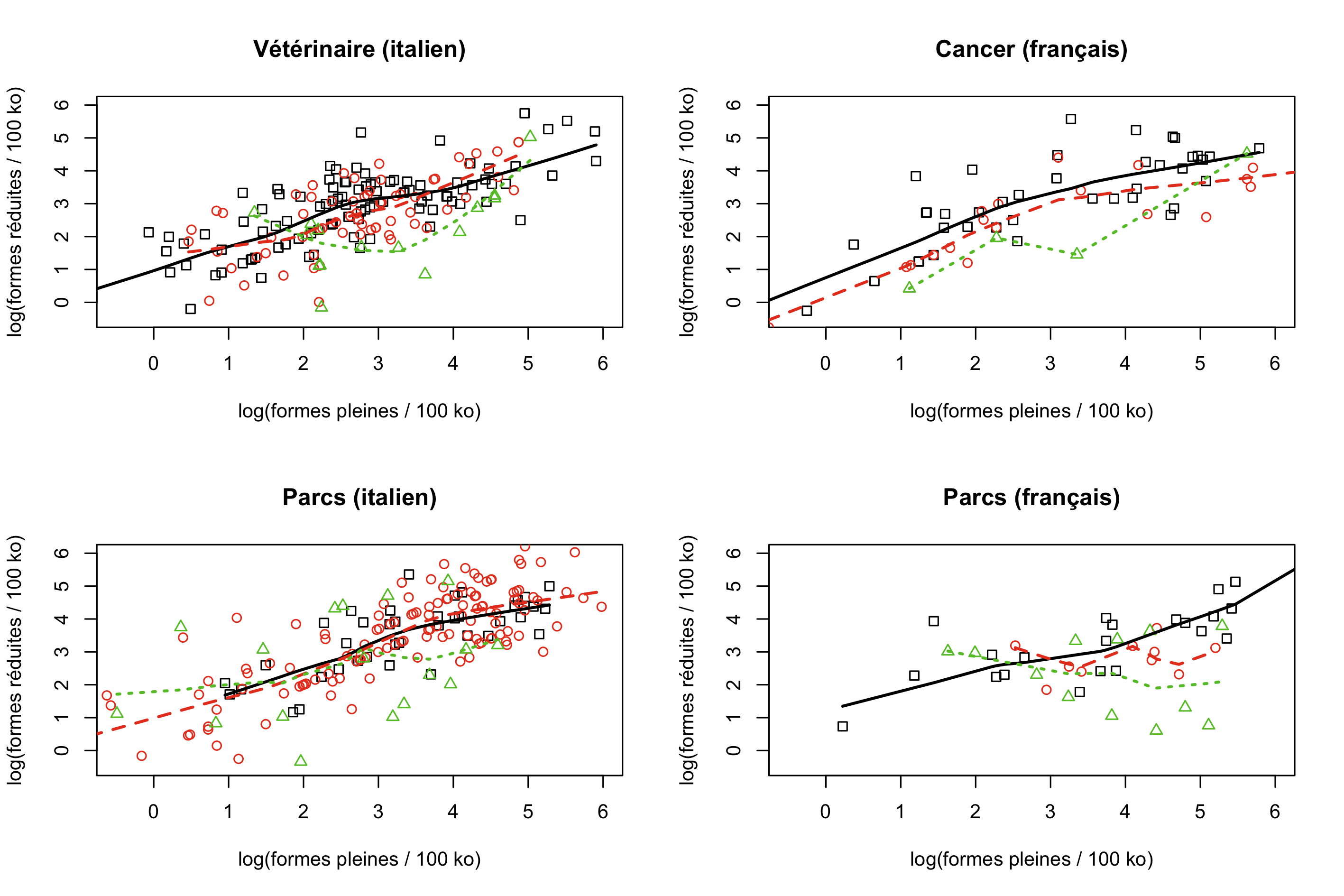}}
\caption{Distribution des documents selon le rapport formes pleines / formes réduites. Les symboles dénotent les trois catégories de texte : carré = cat. 1, cercle = cat. 2, triangle = cat. 3. Les courbes sont des régressions polynomiales LOWESS : courbe continue = cat. 1, tiretée = cat. 2, pointillée = cat. 3.\label{distribs-ana}}
\end{figure}

Dans la suite, nous allons raisonner au niveau des documents.

La fig. \ref{distribs-ana} montre la distribution des documents selon le rapport formes réduites / formes pleines. Les courbes tracées sont des régressions polynomiales des points, selon l'algorithme LOWESS \cite{Cleveland1981}. Elles confirment le fait que l'on a plus de réductions anaphoriques en catégorie 1 (vulgarisation) qu'en catégorie 2 (scient. pédag.) que 3 (scient. spécialisé).

Nous constatons également une différence entre les textes français et italiens. Dans le premier cas on remarque un meilleur regroupement : il y a moins de dispersion pour les cat. 1 et 2 et la cat. 3 se trouve plutôt \emph{sous} la courbe de la cat. 1. Dans le cas italien, il y a beaucoup plus de dispersion et aucun regroupement n'est possible. On peut en déduire que la langue italienne est terminologiquement moins stable que la langue française :  les termes italiens sont moins figés et leur comportement est plus chaotique, l'alternance entre forme complète et forme réduite a tendance à ne pas suivre un schéma régulier. En français, par contre, on trouve que les termes sont plus réguliers.

\subsubsection{Troisième hypothèse} Notre troisième hypothèse peut être reformulée de la manière suivante : \emph{il existe moins de documents avec au moins une réduction cataphorique que de documents avec au moins une réduction anaphorique mais sans aucune réduction cataphorique.}

Cette version de l'hypothèse se vérifie aisément à partir de nos données (toutes catégories de texte confondues), cf. tableau~\ref{troishyp}.
\begin{table}[ht]\small
\begin{center}
\begin{tabular}{|l|c|c|c|c|}\hline
Domaine & Langue & RA & RCA\\\hline
Vétérinaire & italien & 37,47\% & 7,00\%\\
Cancer & français & 11,9\% & 1,49\%\\
Parcs & italien & 48,48\% & 4,71\%\\
Parcs & français & 40,95\% & 4,76\%\\\hline
\end{tabular}\end{center}
\caption{Présence de réductions anaphoriques et cataphoriques dans les documents. La colonne RA (resp. RCA) représente les pourcentages de documents avec au moins une réduction anaphorique (resp. cataphorique).\label{troishyp}}
\end{table}

\section{La réduction lexicale}

\subsection{Hypothèses}

Nous formulons les hypothèses suivantes :

\begin{enumerate}
\item[\rm(1)] \emph{La réduction lexicale dépend du domaine, et est plus fréquente dans les domaines techniques à évolution rapide.}
\item[\rm(2)] \emph{Elle résulte de l'inutilité progressive d'un composant et peut aboutir à une lexicalisation à part entière de la forme réduite. Dans ce cas, on peut observer un \og cycle de vie\fg{} de la forme réduite : naissance, mise en concurrence avec la forme pleine, remplacement de la forme pleine.}
\end{enumerate}

Pour le point (2), nous nous sommes inspirés des articles de Dury \& Drouin sur la nécrologie des termes \cite{Dury2010} et de M{\o}ller \cite{Moller1998} sur la terminochronie.

Dans cette section, nous appelerons \og termes 3-complexes\fg{}, les termes dont les composants sont formés de plus de deux mots.

\subsection{Validation expérimentale}

Contrairement à la réduction anaphorique qui est interne au texte, la réduction lexicale est un mécanisme discursif. Il est donc important d'utiliser des corpus thématiquement stables pour l'étudier. D'ailleurs, la confirmation ou infirmation de la deuxième hypothèse nécessite une datation précise des données textuelles. Ainsi, plutôt que de récupérer des documents arbitraires du Web contenant la forme pleine et des éventuelles formes réduites, nous avons choisi de travailler sur des corpus plus \og conventionnels\fg{} : nous avons étudié le contenu textuel des six revues spécialisées, dans les domaines suivants : les techniques d'emballage, les parcs nationaux et la philosophie (qui nous servira de contre-exemple puisqu'elle est l'antithèse-même d'un domaine \og technique à évolution rapide\fg{}).

Le lecteur trouvera dans le tableau~\ref{listcomp} la liste comparative des revues\footnote{\emph{Kl\=esis} \url{http://www.revue-klesis.org/}, ISSN 1954-3050 ; \emph{Dialegesthai}
\url{http://mondodomani.org/dialegesthai/}, ISSN 1128-5478 ;
\emph{Espaces naturels} \url{http://www.espaces-naturels.fr/a_la_une/la_revue_espaces_naturels}, ISSN 1637-9896 ;
\emph{Parchi} \url{http://www.parks.it/federparchi/rivista/}, sans ISSN ;
\emph{Emballages Magazine} \url{http://www.industrie.com/emballage/}, ISSN 0013-6573 ;
\emph{Italia Imballaggio} \url{http://www.italiaimballaggio.it/italiaimballaggio/05_00/index.html}, ISSN 2037-2183.} considérées.
\begin{table}\small
\begin{center}
\begin{tabular}{|r|c|c|c|c|}\hline
Revue&Domaine&Langue&Années&Volume\\\hline
\emph{Kl\=esis}&philosophie&français&2006--2010&7,25 Mo\\
\emph{Dialegesthai}&philosophie&italien&1999--2010&19,43 Mo\\
\emph{Emballages Magazine}&emballages&français&2002--2009&\ 1,16 Mo\\
\emph{Italia Imballaggio}&emballages&italien&2002--2007&14,05 Mo\\
\emph{Espaces naturels}&parcs&français&2003--2009&\ 5,49 Mo\\
\emph{Parchi}&parcs&italien&1990--2009&\phantom{,49}18 Mo\\\hline
\end{tabular}
\end{center}
\caption{Liste comparative des revues utilisées pour l'étude de la réduction lexicale.\label{listcomp}}
\end{table}
Voici les étapes du protocole utilisé :
\begin{enumerate}
\item extraction des termes 3-complexes{} à l'aide du logiciel Acabit (cf. \S\,\ref{acabit}) ;
\item obtention des formes réduites lexicales potentielles par transformation des formes pleines (A B C $\to$ A C, A B C D $\to$ A B D, A B C D $\to$ A C D, etc.) ;
\item analyse de la distribution des occurrences des formes pleines du (1) et des formes réduites lexicales du (2) dans le corpus, en tenant compte de leur datation et de leurs positions dans les documents.
\end{enumerate}

\subsubsection{Extraction des termes 3-complexes}

Le nombre de termes 3-complexes{} varie énormément d'un corpus à l'autre. Nous n'avons pas tenu compte des termes complexes de la forme \og nom adverbe adjectif(s)\fg{}. Ceux-ci ne sont pas pertinents pour notre étude puisque l'adverbe en modifie le sens. Ainsi, la variation qui consiste à omettre  l'adverbe ne constitue pas une forme réduite lexicale du terme, puisque elle ne renvoie pas au même concept. 

Le nombre de termes 3-complexes{} ainsi obtenus, trié par corpus, est donné dans le tableau~\ref{termesobtenus}.
\begin{table}[ht]\small
\begin{center}
\begin{tabular}{|r|c|c|c|c|}\hline
Revue&$t$&$r$&Occurrences $t$&Occurrences $r$\\\hline
\emph{Kl\=esis}&2&3&3&72\\
\emph{Dialegesthai}&6&6&12&293\\
\emph{Emballages Magazine}&33&15&93&111\\
\emph{Italia Imballaggio}&127&96&218&2\,015\\
\emph{Espaces naturels}&51&45&209&1\,576\\
\emph{Parchi}&88&52&1\,829&24\,953\\\hline
\end{tabular}
\end{center}
\caption{Termes 3-complexes obtenus selon le corpus. $t$, $r$ dénotent, resp., le nombre de termes complexes distincts obtenus (étape 1) et celui de formes réduites (étape 2) attestées dans le corpus. D'autre part \og Occurrences $t$\fg{} (resp. \og Occurrences $r$\fg{}) dénotent le nombre d'occurrences de formes pleines (resp. réduites) dans le corpus.\label{termesobtenus}}
\end{table}
Nous constatons dans le tableau~\ref{termesobtenus} que le nombre de formes pleines et réduites n'est pas directement corrélé avec le nombre d'occurrences de celles-ci dans le corpus : ainsi, dans \emph{Parchi}, on a 24\,953 occurrences des 52 formes réduites attestées, alors que dans \emph{Italia Imballaggio} on a presque deux fois plus de formes réduites attestées (96), avec douze fois moins d'occurrences (2\,015).

\begin{figure}[ht]
\resizebox{1.03\textwidth}{!}{\includegraphics{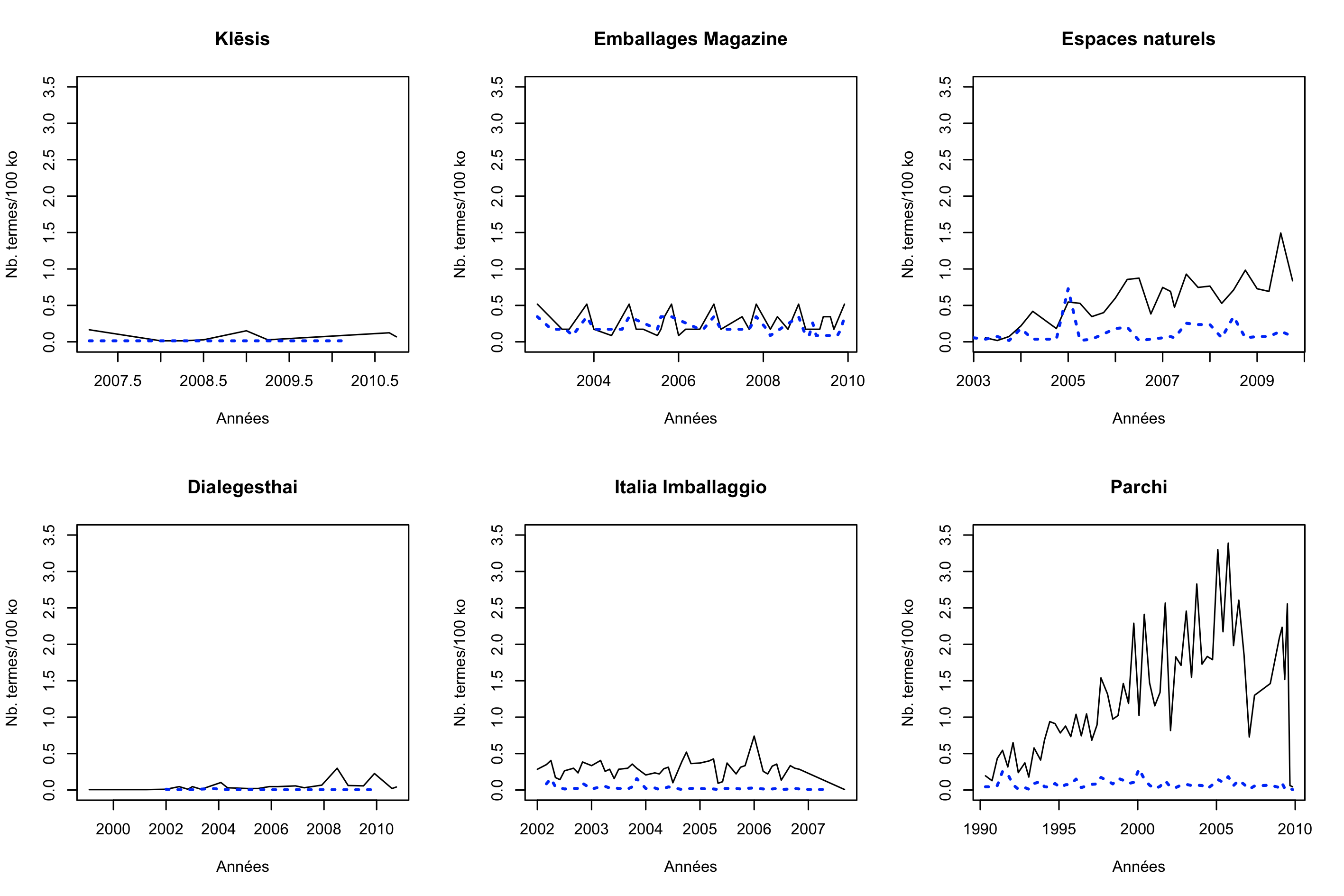}}
\caption{Densité des formes pleines (courbe pointillée) et réduites (courbe  continue) en fonction du temps.\label{distribs}}
\end{figure}

Sur la fig. \ref{distribs} on présente la densité des formes pleines (courbe bleue pointillée) et réduites (courbe continue) en fonction du temps. On voit que les formes réduites sont bien plus nombreuses que les formes pleines, phénomène dû à leur économicité. D'ailleurs dans les cas de \emph{Espaces naturels} et \emph{Parchi} on assiste à une croissance quasi-constante de la densité de formes réduites et à une décroissance du nombre de formes pleines, au fil du temps.

Ces exemples montrent que la réduction lexicale est corrélée à la technicité du domaine, ce qui prouve la première hypothèse.

\subsubsection{Deuxième hypothèse, cycle de vie de terme}

Selon \cite{Dury2010} et \cite{Moller1998}, les termes complexes ont un cycle de vie lié à celui de l'objet qu'ils dénomment. Lorsque l'objet est encore peu connu, la forme pleine est indispensable pour le dénommer. Mais dans les cas où, progressivement, l'objet se répand, les réductions lexicales deviennent possibles, puisque le risque d'ambiguïté est moindre. Enfin, dans les cas où certains constituants de la forme pleine n'ont aucune charge sémantique réelle, la réduction lexicale finit par remplacer complètement la forme pleine.

Pour déceler des traces du phénomène de \og cycle de vie de terme\fg{}, il a d'abord fallu \og unifier\fg{} les mesures de datation et de position dans le texte. Pour cela, nous avons suivi l'approche suivante : 
\begin{enumerate}
\item nous définissons un corpus $\mathcal{C}$ comme étant une suite de $N$ documents $(D_i)_{1\leq i\leq N}$ datés. La datation est une fonction strictement croissante $T:\mathcal{C}\to\mathbb{Q}^N$ où $T(D_i)=\mathrm{ann\acute{e}e}(D_i)+\frac{\mathrm{mois}(D_i)-1}{12}$ ;
\item en considérant un document $D_i$ comme une suite de caractères $(c_{i,j})$ nous définissons la \emph{fonction de datation généralisée} $T^*$ comme la fonction linéaire par morceaux qui étend $T$ de au niveau des caractères :
$$
\xymatrix{(D_*) \ar[r]^T \ar[d]^{\pi} & \mathbb{Q}^N\\
(c_{*,*}) \ar[ru]_{T^*} & {}}
$$
où $\pi$ est la projection qui envoie le document $D_i$ sur son premier caractère $c_{i,1}$. 

Autrement dit, si le document $D_5$ date du 1\up{er} mai 2005 et $D_6$ du 1\up{er} juin de la même année, alors $T(D_5)=2005 + \frac{5-1}{12}=2005,333$, $T(D_6)=2005,417$ ; si une occurrence de terme se trouve à la position $37\,238$ de $D_i$ (dont la taille est, par exemple, de $57\,642$ caractères) alors sa valeur de datation généralisée sera $T^*(t_i)=T(D_5)+\frac{37\,238}{57\,642}\cdot(T(D_6)-T(D_5))=2\,005,3873$. Ainsi toute occurrence a une valeur de datation unique, compatible avec la datation des fichiers et avec l'ordre linéaire du texte\footnote{Suite à la remarque d'un relecteur de l'article, notons que cette fonction convient à des corpus figés. En effet, dans le cas d'un corpus dynamique, l'ajout de documents supplémentaires changerait les valeurs obtenues pour les anciens documents.} ;

\item la fonction de datation généralisée nous permet de définir une distance temporelle $d_T$ entre les occurrences de termes dans le corpus tout entier : $d_T(t_i,t_j)=T^*(t_j)-T^*(t_i)$ ;
\item ainsi, pour un terme donné, les occurrences de formes pleines $p_*$ et de formes réduites $r_*$ deviennent des ensembles ordonnés de points de la droite temporelle.
\end{enumerate}

La modélisation étant faite, posons-nous le problème de représentation du phénomène de \og cycle de vie de terme\fg{}. On ne peut prendre simplement les premières occurrences de formes pleine et réduite, puisque celles-ci peuvent très bien être des \emph{outliers}. Prendre la moyenne arithmétique des valeurs temporelles de toutes les formes pleines (resp. réduites) ne serait pas une solution non plus, puisqu'on s'intéresse aux débuts de l'apparition d'une forme et non pas à son historique complet. Il convient de traiter séparément les cas où le nombre d'occurrences prend de l'ampleur et ceux où il reste limité. Dans le premier cas, on limitera le nombre d'occurrences à considérer, dans le deuxième cas on prendra la totalité des occurrences.

Comme nous voulons éviter l'aberration causée par des éventuels \emph{outliers}, tout en gardant un maximum d'information pour le cas où le nombre d'occurrences est très limité, nous avons choisi de considérer la \emph{moyenne géométrique des $N$ premières occurrences}. En effet, la moyenne géométrique est mieux indiquée pour diminuer l'impact des \emph{outliers} que la moyenne arithmétique. Dans les résultats présentés ci-dessous, on a pris $N=100$.

\begin{figure}[ht]
\resizebox{1.03\textwidth}{!}{\includegraphics{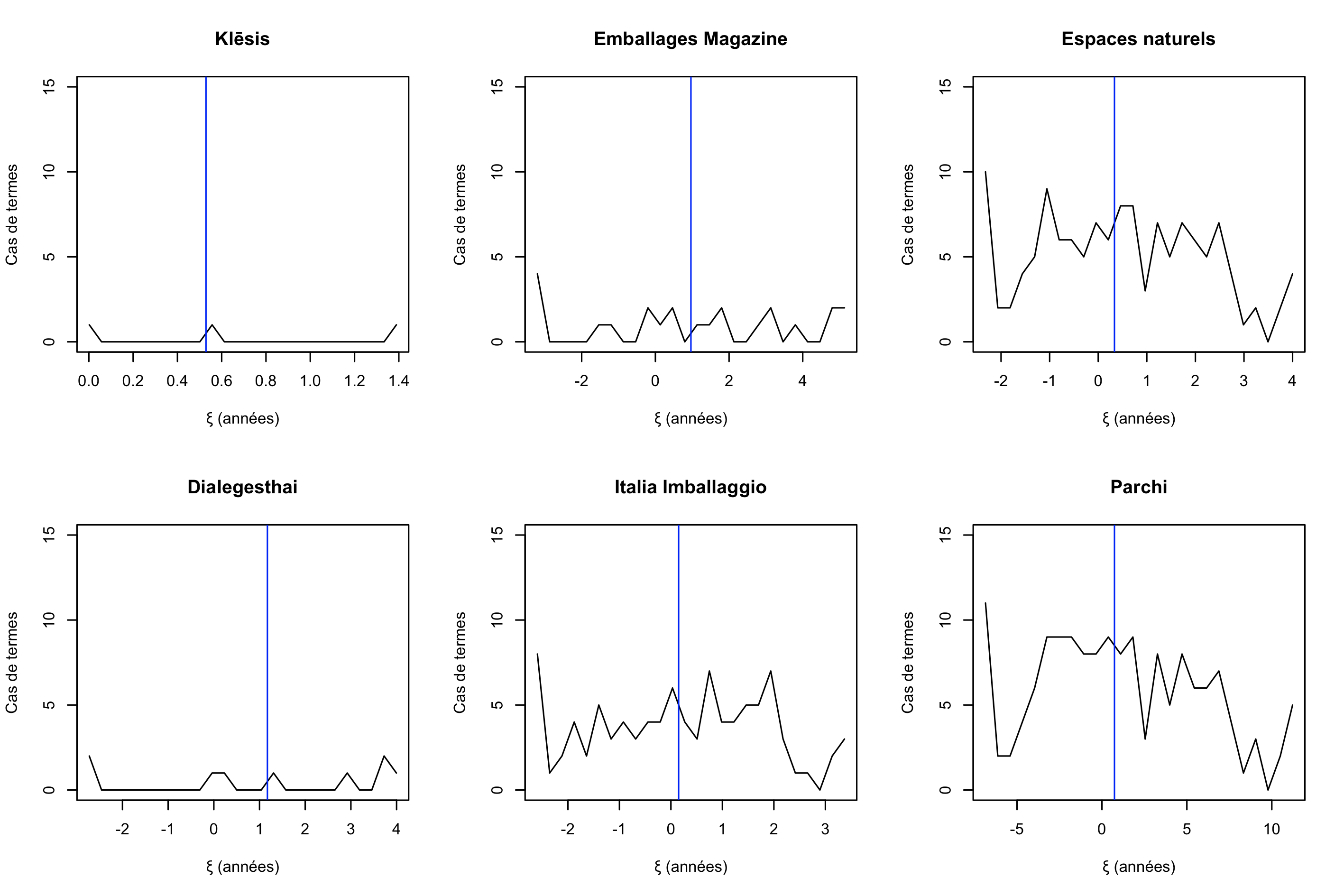}}
\caption{Densité des valeurs de $\xi$. La ligne bleue indique l'emplacement de la médiane.\label{delta}}
\end{figure}

Nous allons donc calculer, pour chaque terme de chaque corpus, la quantité $\xi=\bar{r}-\bar{t}$, où $\bar{r}$ et $\bar{t}$ sont les moyennes géométriques des 100 premières occurrences des formes réduites (resp. de la forme pleine) du terme. La fig. \ref{delta} représente la densité de $\xi$. On constate que la médiane de cette densité est positive dans tous les corpus, ce qui valide notre deuxième hypothèse.

\section{Conclusions et perspectives}\label{conclusions}

En traitant des corpus dans divers domaines et dans les deux langues (français et italien) nous avons validé expérimentalement nos hypothèses : que la réduction anaphorique dépend du type de discours, mais ne dépend pas du domaine ou de la langue ; que les formes réduites anaphoriques ont tendance à suivre les formes pleines, plutôt que de les précéder ; que la réduction lexicale dépend du domaine, et est plus fréquente dans les domaines techniques à évolution rapide ; et que les formes pleines suivent un cycle de vie et sont souvent remplacées par des formes réduites lexicales.

Les logiciels de type \emph{Acabit} \cite{Daille1996} ou \emph{TerMine} \cite{Ananiadou1998} extraient efficacement des termes complexes à partir de données textuelles en langue de spécialité, mais ne nous fournissent aucune indication sur les éventuelles relations sémantiques entre eux. Et pourtant, dans les langues de spécialité les relations d'hyponymie/hypéronymie entre les termes sont d'une importance capitale.

Dans ce travail nous avons essayé de fournir quelques indications sur la probabilité d'avoir, dans certains cas que nous nous proposons de décrire, des relations (quasi-)synonymiques.

Un terme complexe peut être décrit en tant que $tc_1\ldots c_n$ où $t$ est la tête et $c_i$ les composants. Prenons, par exemple, $n=3$. En réduisant le terme $tc_1c_2c_3$ nous obtenons le treillis suivant (où les flèches sont des réductions purement formelles, sans aucune hypothèse sur les relations semantiques entre les termes) :
$$
\xymatrix{{}&tc_1c_2c_3\ar@{->}[dl]^1\ar@{->}[d]^2\ar@{->}[dr]^3&{}\\
tc_1c_2\ar@{->}[d]^4\ar@{->}[dr]^5&tc_1c_3\ar@{->}[dl]^6\ar@{->}[dr]^7&tc_2c_3\ar@{->}[dl]^8\ar@{->}[d]^9\\
tc_1\ar@{.>}[dr]^{10}&tc_2\ar@{.>}[d]^{11}&tc_3\ar@{.>}[dl]^{12}\\
{}&t&{}}
$$
Selon les définitions données dans la première partie de cet article, si réduction anaphorique (ou cataphorique) il y a, elle correspond forcément aux flèches 1, 4 et 10, c'est-à-dire le bord gauche du treillis. Les autres flèches du treillis peuvent, par contre, être des réductions lexicales.

Les hypothèses que nous avons démontré experimentalement nous ont conduit à formuler les règles suivantes :
\begin{enumerate}
\item si une réduction est du type $tc_1\ldots c_n\to tc_1\ldots c_{n-1}$, si la forme pleine $tc_1\ldots c_n$ apparaît avant la forme réduite $tc_1\ldots c_{n-1}$, et si le texte est de catégorie~1 (vulgarisation), alors il y a des fortes chances que la réduction soit anaphorique.

Dans ce cas, s'agissant d'une anaphore, les mécanismes habituels de résolution d'anaphore peuvent être utilisés pour contribuer à la confirmation ou infirmation de l'hypothèse de réduction anaphorique ;

\item pour le même type de réduction, si la forme pleine apparaît après la forme réduite et/ou si le texte est de catégorie~3 (discours scientifique spécialisé ou texte législatif), alors il y a peu de chances que ce soit une réduction anaphorique ;

\item pour tout autre type de réduction dans le treillis ci-dessus, si le texte appartient à un domaine technique à évolution rapide, la possibilité d'une réduction lexicale (due à l'inutilité progressive d'un composant) existe. Pour la confirmer ou l'infirmer, il faudra utiliser des méthodes sémantiques. On pourra, par exemple, former des vecteurs de concepts environnant les deux termes, dont les coefficients seraient pondérés par la fréquence d'occurrences au niveau du corpus. En comparant les vecteurs (par exemple, en calculant leur cosinus), on aurait une indication plus forte sur une éventuelle synonymie.
\end{enumerate}

Partant de là, nombreuses sont les pistes qui peuvent être suivies :
\begin{enumerate}
\item en se basant sur notre corpus annoté, se poser la question de la pertinence de l'application des algorithmes traditionnels de résolution d'anaphore pour confirmer ou infirmer la réduction anaphorique ;

\item idem, pour les mesures de similarité sémantique et la réduction lexicale. Notons que si la résolution d'anaphore et la mesure de similarité s'avèrent pertinentes dans ce domaine, celles-ci étant des techniques lourdes, notre approche peut fournir des critères pour évaluer la pertinence de leur utilisation dans un texte ;

\item étendre notre étude à d'autres langues comme l'anglais (dont les mécanismes de réduction lexicale sont bien différents), l'allemand (qui fourmille de mots composés), les langues idéographiques (où le caractère est porteur de sens et où on assiste à un triangle terme - caractère - concept), etc. ; 

\item étudier l'interaction des réductions lexicales et anaphoriques : en effet, nous avons constaté (sans en tenir compte dans nos calculs) le phénomène suivant : une forme pleine (avec arbre anaphorique) était suivie d'une réduction lexicale possédant son propre arbre anaphorique mais dont les feuilles représentaient les mêmes formes que celles des feuilles de l'arbre de la forme pleine. Autrement dit, on peut qu'une forme réduite anaphorique se réfère à une forme pleine ou à une réduction lexicale de celle-ci. Cela nous incite à fusionner les deux arbres (ceux des formes pleine et réduite lexicale) en un seul graphe qui puisse modéliser les deux types de réduction ;

\item étudier la correlation des différentes quantités définies dans cet article avec la $C$-valeur de Frantzi-Ananiadou-Tsujii \cite{Ananiadou1998} ;

\item détecter les éventuels marqueurs anaphoriques dans le voisinage des formes réduites anaphoriques et se servir d'algorithmes de résolution d'anaphore pour obtenir un modèle plus riche de l'arbre anaphorique, dans lequel les arêtes représenteront, dans la mesure du possible, le lien entre anaphore et antécédent ;

\item si l'\og indice de réductibilité\fg{} est la probabilité qu'un terme complexe soit la réduction (lexicale ou anaphorique) d'un autre terme complexe, développer un algorithme qui calcule cet indice en se basant sur différents facteurs (type de texte, domaine, position dans le texte, etc.). Cette information serait d'une grande utilité pour l'analyse sémantique, l'indexation ou la traduction automatique.
\end{enumerate}

\bibliography{article}

\end{document}